\documentclass{WileyMSP-template}
\usepackage{subcaption}
\usepackage{wrapfig}
\usepackage{amsmath}
\usepackage{amsfonts}

\usepackage[hidelinks]{hyperref}

\begin{document}

\pagestyle{fancy}
\rhead{\includegraphics[width=2.5cm]{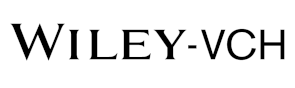}}

\title{Cost-Effective Radar Sensors for Field-Based Water Level Monitoring with Sub-Centimeter Accuracy}

\maketitle

\author{Anna Zavei-Boroda}
\author{J. Toby Minear}
\author{Kyle Harlow}
\author{Dusty Woods}
\author{Christoffer Heckman*}

\begin{affiliations}
A. Zavei-Boroda, K. Harlow, D. Woods, C. Heckman\\
Department of Computer Science\\
University of Colorado, Boulder\\
Email Address: \verb+christoffer.heckman@colorado.edu+

J. T. Minear\\
Cooperative Institute for Research in Environmental Science\\
University of Colorado, Boulder

\end{affiliations}

\keywords{Millimeter Wave Radar, Water Level Monitoring}

\begin{abstract}
Water level monitoring is critical for flood management, water resource allocation, and ecological assessment, yet traditional methods remain costly and limited in coverage. This work explores radar-based sensing as a low-cost alternative for water level estimation, leveraging its non-contact nature and robustness to environmental conditions. Commercial radar sensors are evaluated in real-world field tests, applying statistical filtering techniques to improve accuracy. Results show that a single radar sensor can achieve centimeter-scale precision with minimal calibration, making it a practical solution for autonomous water monitoring using drones and robotic platforms.
\end{abstract}

\section{Introduction}

\subsection{Water Level Monitoring}

Accurate monitoring of water levels in lakes, reservoirs, and rivers provides the data necessary for ecological conservation, disaster prevention, and resource management~\cite{intro1book}. Reliable water level data support effective flood mitigation strategies, inform hydroelectric power generation decisions, facilitate agricultural irrigation planning, and preserve ecological stability~\cite{intro2}. Despite their importance, many water bodies remain unmonitored due to limitations in traditional monitoring methods, leading to gaps in environmental and hydrological data essential for informed decision-making and disaster preparedness~\cite{intro3}.

Traditional water level measurement approaches primarily include in-situ gauges, ultrasonic sensors, vision-based optical methods, and satellite remote sensing. These established techniques have varying limitations related to installation complexity, cost, environmental robustness, and data resolution, as summarized in \textbf{Table~\ref{tab:wlm_methods}}. 

\begin{table}
  \caption{Comparison of Water Level Monitoring Methods}
  \label{tab:wlm_methods}
  \begin{tabular}[htbp]{@{}llll@{}}
    \hline
    \textbf{Device Type} & \textbf{Cost} & \textbf{Installation} & \textbf{Limitations} \\
    \hline
    Pressure Transducer & Low & Submerging required & Sensitive to sediment build-up, signal drift over time \\
    Float Gauge & Low & Advanced mechanical & Manual maintenance required \\
    Vision-Based & Low & Above surface & Sensitive to lighting conditions, obstructions, and disturbances \\
    Ultrasonic & Moderate & Above surface & Sensitive to weather conditions \\
    Satellite & High & None & Low resolution, infrequent data updates, high latency \\
    Radar & Low & Above surface & Signal noise, multipath reflections \\
    \hline
  \end{tabular}
\end{table}

\subsection{Radar Sensing}

Integrated sensing approaches have emerged as scalable and cost-effective alternatives for environmental monitoring. These sensing techniques are often due to developments in robotic sensors and algorithms designed for mobile platforms, manipulation, and other applications (e.g., autonomous driving). Robotic platforms equipped with radar, lidar, or optical sensors can autonomously collect large-scale environmental data, significantly improving the frequency and spatial coverage of measurements~\cite{intro4}. Among these sensing modalities, radar is particularly well-suited for water level monitoring because of its unique all-weather resilience and capability to perform non-contact measurements. Unlike optical and laser-based methods, radar sensors remain effective in fog, rain, and visually degraded conditions, making them reliable for long-term deployments in natural environments.

FMCW (frequency modulated continuous wave) mmWave (millimeter-wave) radar is especially advantageous for intelligent perception tasks due to its high precision in measuring both range and velocity of objects \cite{ti2017radar}. An FMCW mmWave radar is a type of radar system that continuously transmits a variable-frequency signal and simultaneously receives echoes of that signal that are offset and scaled due to range, angle of arrival, and Doppler effects. The type of radar used in this work operates in relatively short radio wavelengths (76--81 GHz); however, in general, mmWave radars can operate in frequencies from 30--300 GHz.

These radars, typically used in automotive or field robotics, have multiple transmit antennae (TX) and receive antennae (RX). Each transmit-receive antenna pair forms a ``virtual antenna.'' Radar emissions commonly known as ``chirps'' are composed of frequency modulation patterns, where the frequency increases linearly over a specified duration. By analyzing the frequency shift between the transmitted chirp and its received echo, the radar determines the range and velocity of an object through time-of-flight calculations. There are many details associated with the design of the chirp pattern that are further described in \cite{ti2020chirp}.

To determine the distance (or range) to an object, a single chirp can be used. A transmitted chirp is combined with its received reflection to produce an intermediate frequency (IF) signal. For a simple case of a static object, this new signal has a constant frequency equal to the difference of the frequencies of the transmitted and received chirps. The delay between the transmission of a chirp and registering of its reflection is denoted as $\tau$. Knowing the delay, as well as the starting frequency of the chirp $f_c$, we can calculate the initial phase of the IF signal as:
\begin{equation}
    \phi_0 = 2\pi f_c\tau,
    \label{eq:phase0}
\end{equation}
which is then used to determine the range $d$ as:
\begin{equation}
    d = \frac{\lambda\phi_0}{4\pi}, \label{eq:distance}
\end{equation}
where $\lambda$ is the wavelength of the chirp.

In the general case with multiple chirps, each received chirp is first processed using a range FFT (fast Fourier transform) to detect peaks corresponding to object distances. For a given peak, the phase of the FFT output varies across successive chirps if the object is moving. The phase difference $\Delta\Phi$ between consecutive chirps can be used to estimate the object's radial velocity as shown in Equation~\ref{eq:dopplerv}. To resolve multiple objects located at the same range but moving with different velocities, a Doppler FFT is applied across chirps for each range value, producing a range--Doppler signal array.
\begin{equation}
    v = \frac{\lambda\Delta\Phi}{4\pi T_c} \label{eq:dopplerv}
\end{equation}
where $T_c$ is the chirp's duration.

The post-FFT data can be further processed using the constant false alarm rate (CFAR) algorithm, a fundamental detection method used in radar-based point cloud generation~\cite{cfar}. CFAR operates on the magnitude of the range--Doppler signal array and extracts true signal peaks corresponding to reflections from physical objects by dynamically adjusting the detection threshold based on the local noise or clutter level surrounding each cell. By evaluating each cell against a threshold calculated from its neighboring reference cells, CFAR ensures a consistent probability of false alarm across varying signal conditions.

Radar-based sensing faces unique challenges, particularly when applied to monitoring water surfaces:
\begin{itemize}
    \item \textit{Sensor Noise.} Radar signals inherently contain electronic and environmental noise, which can lead to spurious detections or inaccurate distance estimates.

    \item \textit{Multipath Reflections.} Multipath effects occur when radar signals reflect off multiple surfaces before returning to the sensor, resulting in distorted or ambiguous measurements.

    \item \textit{Water-Specific Challenges.} The physical properties of water surfaces introduce additional complexity. Depending on the radar frequency and water composition, radar waves may partially penetrate the surface, producing reflections from within the water column rather than exclusively from the surface. Variations in turbidity, temperature, and the presence of submerged objects or organisms (e.g., fish) can further affect signal behavior. In addition, dynamic surface phenomena such as waves, ripples, and floating debris contribute to measurement variability.
\end{itemize}

Addressing these challenges is essential for ensuring accurate and reliable measurements in real-world applications. As described in the following sections, a combination of off-the-shelf and novel algorithmic filtering techniques is employed to mitigate these sources of noise and bias in the measured signal.

\subsection{Contribution of This Work}

\begin{wrapfigure}{l}{0.5\textwidth} 
\centering
  \vspace{-0.1em}
  \hspace{-5.5em}
  \begin{minipage}{0.25\linewidth}
    \includegraphics[width=1.2\linewidth]{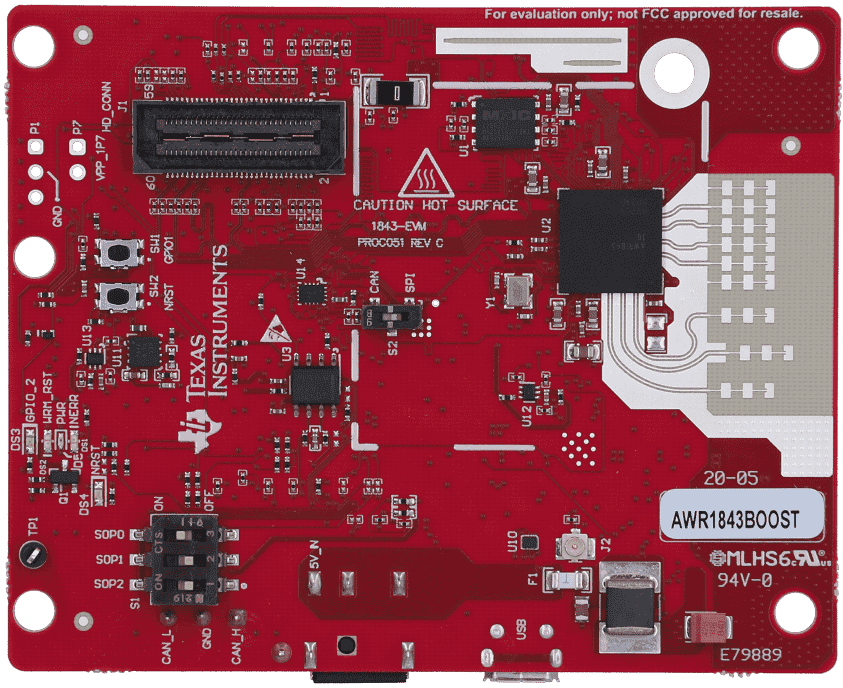}
    \vspace{-1.5em}
    \caption*{(a) AWR1843}
    \vspace{0.3em}
    \includegraphics[width=1.2\linewidth]{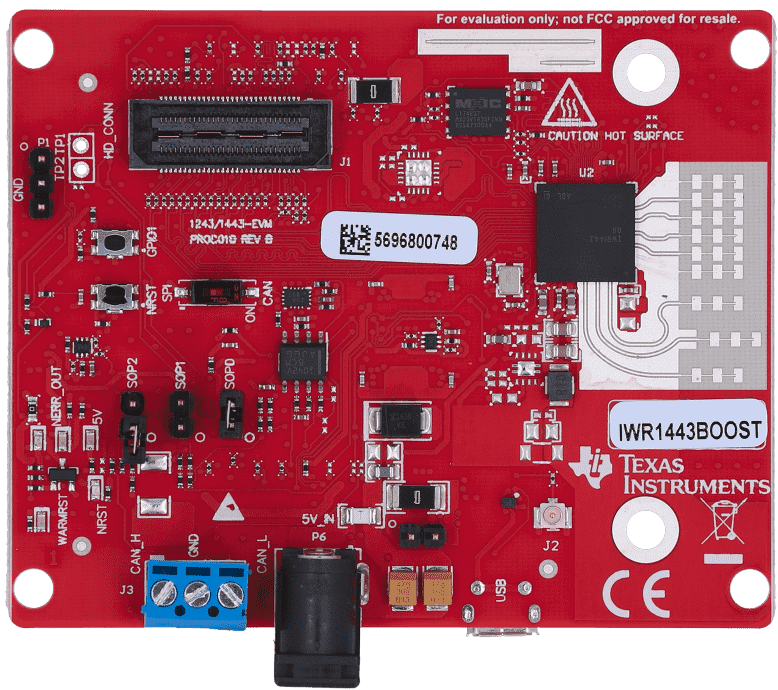}
    \vspace{-1.5em}
    \caption*{(b) IWR1443}
  \end{minipage}
  \hspace{0.8em}
  \begin{minipage}{0.45\linewidth}
    \includegraphics[width=1.5\linewidth]{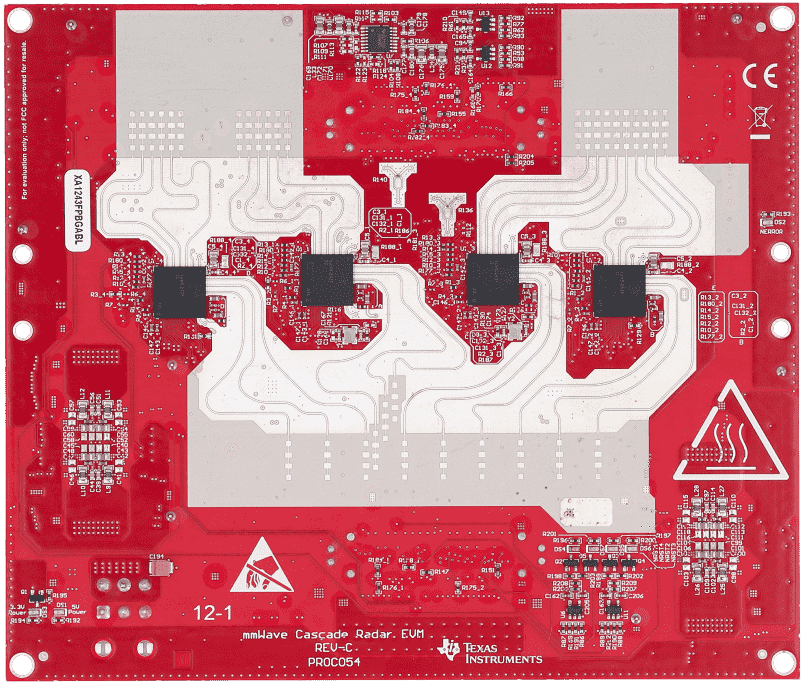}
    \caption*{(c) MMWCAS-RF Cascade}
  \end{minipage}

  \caption{Radar sensors evaluated in this work.}
  \label{fig:radar_sensors}
  \vspace{-1em}
\end{wrapfigure}

This work evaluates commercially available radar sensors for water level monitoring in real-world environments, with experimental results demonstrating sub-centimeter root-mean-square error (RMSE). Three radar sensors from Texas Instruments (TI) are selected for evaluation: (a) the TI AWR1843, (b) the TI IWR1443, and (c) the TI MMWCAS-RF Cascade, as shown in Figure~\ref{fig:radar_sensors}. The AWR1843 and IWR1443 are compact, single-chip mmWave radar sensors, while the Cascade system is a cascaded multi-chip radar platform providing a substantially larger virtual antenna array. For the AWR1843 and Cascade sensors, manufacturer-provided default waveform configurations are employed, whereas a fluid-specific sensing waveform configuration~\cite{ti2025rwlm} is used for the IWR1443. All sensor configurations, collected datasets, and processing code are publicly available~\cite{zenodo}.

This work makes primary contributions in the development of filtering and water level estimation algorithms that operate on the output of these sensors. The sensors are assessed for their water-sensing capabilities through multiple outdoor field experiments. The sensor readings are processed with parametrized filtering techniques to detect the distance from the sensor to the water surface accurately while accounting for the radar-specific challenges. The experimental results highlight the capability of radar sensors to consistently achieve centimeter-scale accuracy with minimal calibration, demonstrating their feasibility for automated water level monitoring. Accessible radar-based water level monitoring enables integration with robotic platforms, such as drones and unmanned surface vehicles, for autonomous collection of environmental data across extensive and inaccessible regions.

\section{Related Work}
Radar sensing has been widely used in robotic perception, particularly for navigation, mapping, and environmental monitoring~\cite{harlow2024new}. Unlike optical sensors, radar provides long-range, non-contact sensing and has been integrated into autonomous systems for terrain mapping, object detection, and localization~\cite{ajay2023rmap, kramer2021coloradar}. Recent advancements have improved the use of millimeter-wave radar in robotics, enabling high-resolution environmental perception and multi-sensor fusion~\cite{harlow2024new}.

In hydrological monitoring, radar-based sensing has been applied to river flow analysis, flood prediction, and hydrological modeling. Prior studies have demonstrated the use of FMCW radar to analyze river surface dynamics by extracting characteristic flow patterns~\cite{rw1}. Other work has focused on radar-based real-time flood forecasting, using rainfall data to predict urban water level fluctuations~\cite{rw2}. Doppler radar sensors have also been deployed to estimate river surface velocity and discharge as a non-contact hydrological measurement approach~\cite{rw3}. In contrast, satellite-based techniques, such as L-band InSAR for coastal wetland monitoring, enable large-scale hydrological assessment but are limited by coarse temporal resolution and data availability~\cite{rw4}.

\begin{wrapfigure}{tr}{0.25\textwidth}
    \centering
    \vspace{-1em}
    \includegraphics[width=\linewidth]{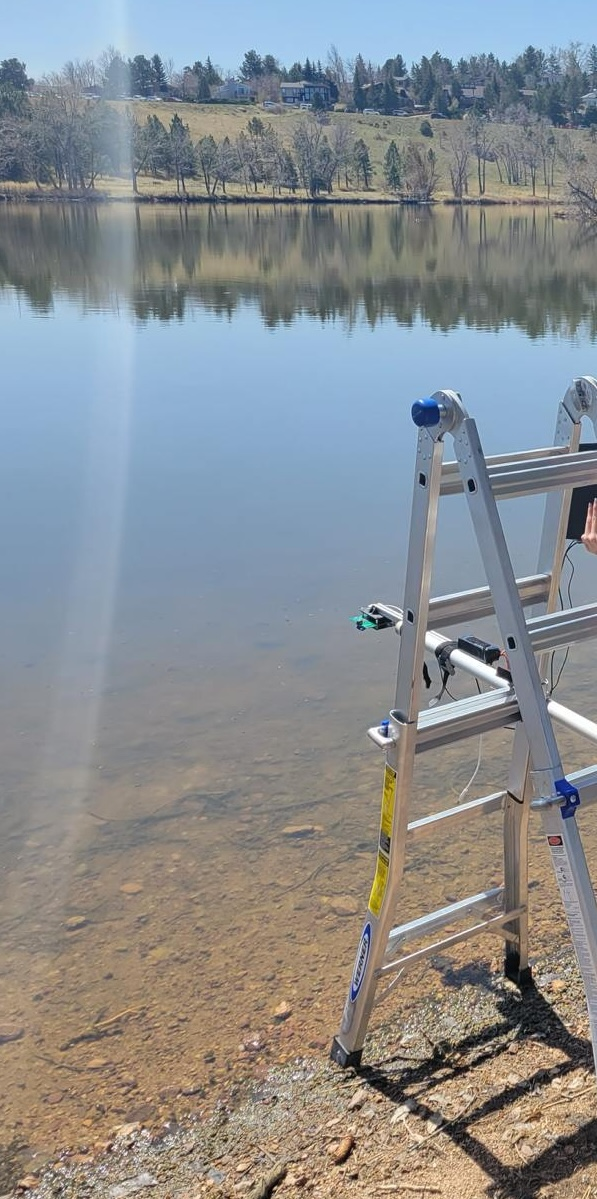}
    \caption{Experimental setup.}
    \label{fig:rig_real}
    \vspace{-3em}
\end{wrapfigure}

This work builds upon recent advancements in robotic radar perception, particularly in the development of millimeter-wave radar sensing for environmental applications. It focuses on low-cost, commercially available radar sensors for static water level monitoring, emphasizing real-world field deployment, ease of use, and cost-effectiveness~\cite{rl1, rl2}. This approach enables broader adoption of radar-based environmental monitoring without requiring complex calibration procedures or expensive infrastructure.

\subsection{Experimental Setup}

The goal of the experiments conducted in this work is to provide insight into large-scale data collection missions through small-scale analogues. In this section and throughout the paper, the term \textit{deployment} refers to a single outdoor field experiment. Each deployment consists of one or more continuous recordings, referred to as \textit{runs}. A \textit{manual deployment} is a short field experiment in which the sensing rig is started and stopped manually to collect sensor readings. During a manual deployment, the groundtruth water level is estimated manually using a marked pole. In contrast, an \textit{automated deployment} is an experiment in which the setup is started and stopped autonomously, enabling stable and periodic runs over an extended period of time. In this case, the water level is validated using pre-installed pressure transducer measurements. Since a single run typically corresponds to a recording lasting a few minutes, manual deployments contain only a small number of runs, whereas an automated deployment allows days’ worth of data to be recorded from the same water body.

\section{Methodology}

\subsection{Experimental Setup}

The goal of the experiments conducted in this work is to provide insight into large-scale data collection missions through small-scale analogues. In this section and below, the word \textit{deployment} refers to a single outdoor field experiment. Each deployment consists of one or more continuous recordings, referred to as \textit{runs}. A \textit{manual deployment} is a short field experiment in which the sensing rig is started and stopped manually to collect sensor readings. During a manual deployment, the groundtruth water level is estimated manually using a marked pole. In contrast, an \textit{automated deployment} is an experiment in which the setup is started and stopped autonomously, enabling stable and periodic runs over an extended period of time. In this case, the water level is validated using pre-installed pressure transducer measurements. Since a single run typically corresponds to a recording lasting a few minutes, manual deployments contain only a small number of runs, whereas an automated deployment allows days’ worth of data to be recorded from the same water body.

For each deployment, the radar sensors were mounted on a plank supported by a ladder (\textbf{Figure~\ref{fig:rig_real}}) at a fixed distance above the water surface, oriented downward. The setup tolerated sensor tilts up to 30 degrees, compensated by an onboard 6-axis inertial measurement unit (IMU). The radar sensor is powered by a battery and connected to a sensor control and data collection computer. For an automated deployment, both the computer and the sensor are placed in a transparent case for protection.

\begin{wrapfigure}{l}{0.5\textwidth}
    \centering
    \includegraphics[width=\linewidth]{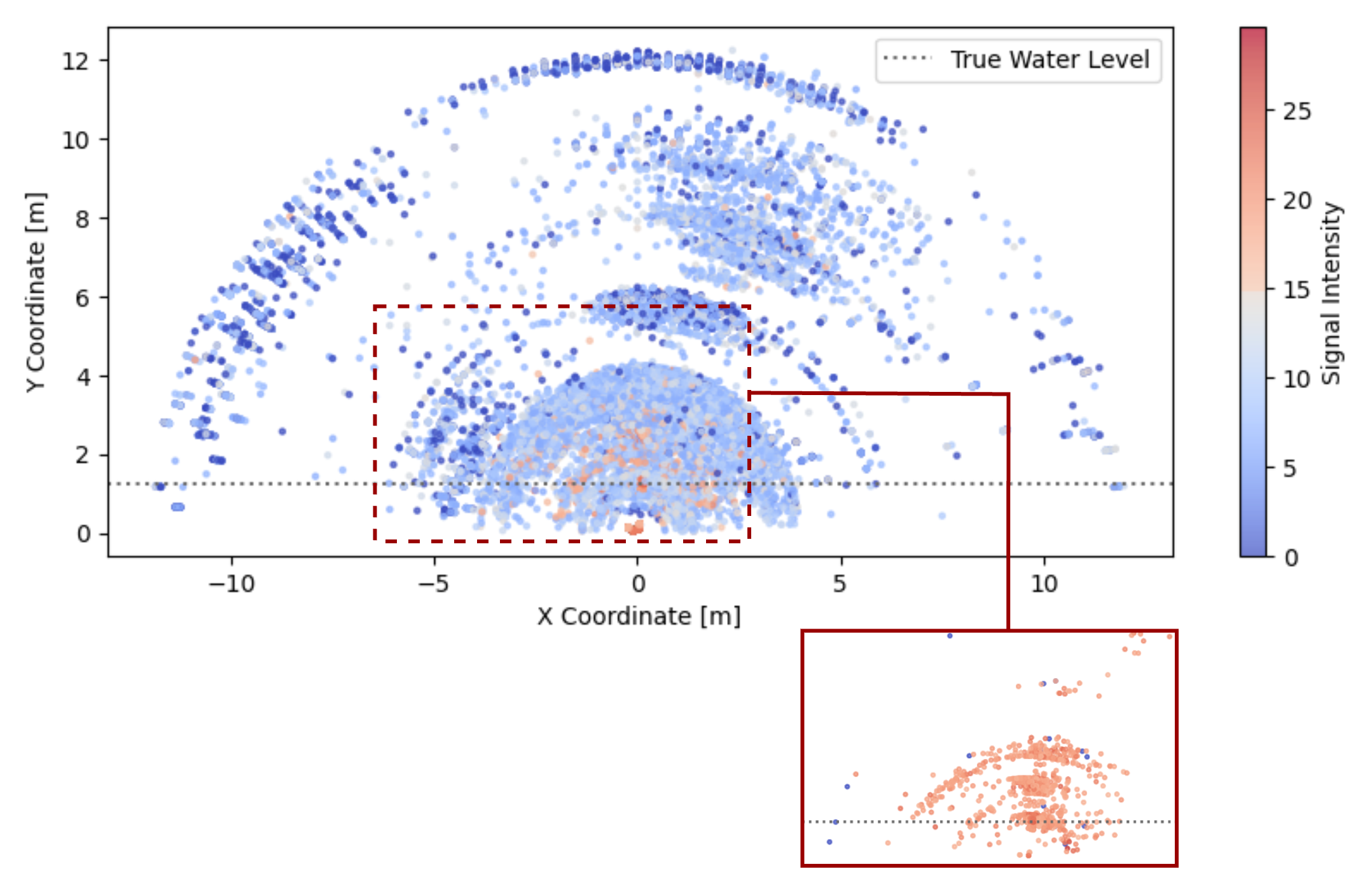}
    \caption{An example of several radar measurements accumulated in one plot as 2D point clouds. The $Y$ axis shows the vertical distance to the point, while the color denotes the radar signal intensity from high (red) to low (blue). Filtering the points by intensity results in a smaller cloud specifically around the water mass under the sensor.}
    \label{fig:cloud}
    \vspace{-2em}
\end{wrapfigure}

To benchmark the basic feasibility of sensing the water surface, a sequence of initial manual deployments was performed. The results below focus on two specific manual deployments that produced useful evidence with regard to selecting the final sensor and data processing approach. The experiments end with an automated 3-day deployment where the final approach is verified under more realistic conditions.

\subsection{Sensing Configurations}

One cloud measurement is defined as a collection of 3D points, each storing the $XYZ$ coordinates of the reflection location and the reflected signal intensity $p$. In this work, the default configurations are used for the AWR1843 and Cascade sensors. Owing to the IWR1443 sensor's waveform configuration specifically designed for fluid sensing, its measurements consist of a single 2D point that is most likely associated with the target fluid, rather than a full 3D point cloud.

The $y$ coordinate of each point represents the distance from the sensor to the reflection and serves as the primary quantity for water level monitoring. The $x$ coordinate is secondary and is used for filtering purposes. For 3D measurements, the $z$ coordinate is collapsed by projecting the points into the $XY$ plane. After recording a sequence of measurements, the resulting $XYZP$ points are filtered to estimate the true water level. Figure~\ref{fig:cloud} shows an example of a 2D point cloud produced by the AWR1843 over several minutes of measurements. Higher-intensity (red) points tend to correspond to reflections from the water mass beneath the sensor, while lower-intensity (blue) points typically lie outside the target region.

\begin{wrapfigure}{r}{0.5\textwidth}
    \centering
    \includegraphics[width=\linewidth]{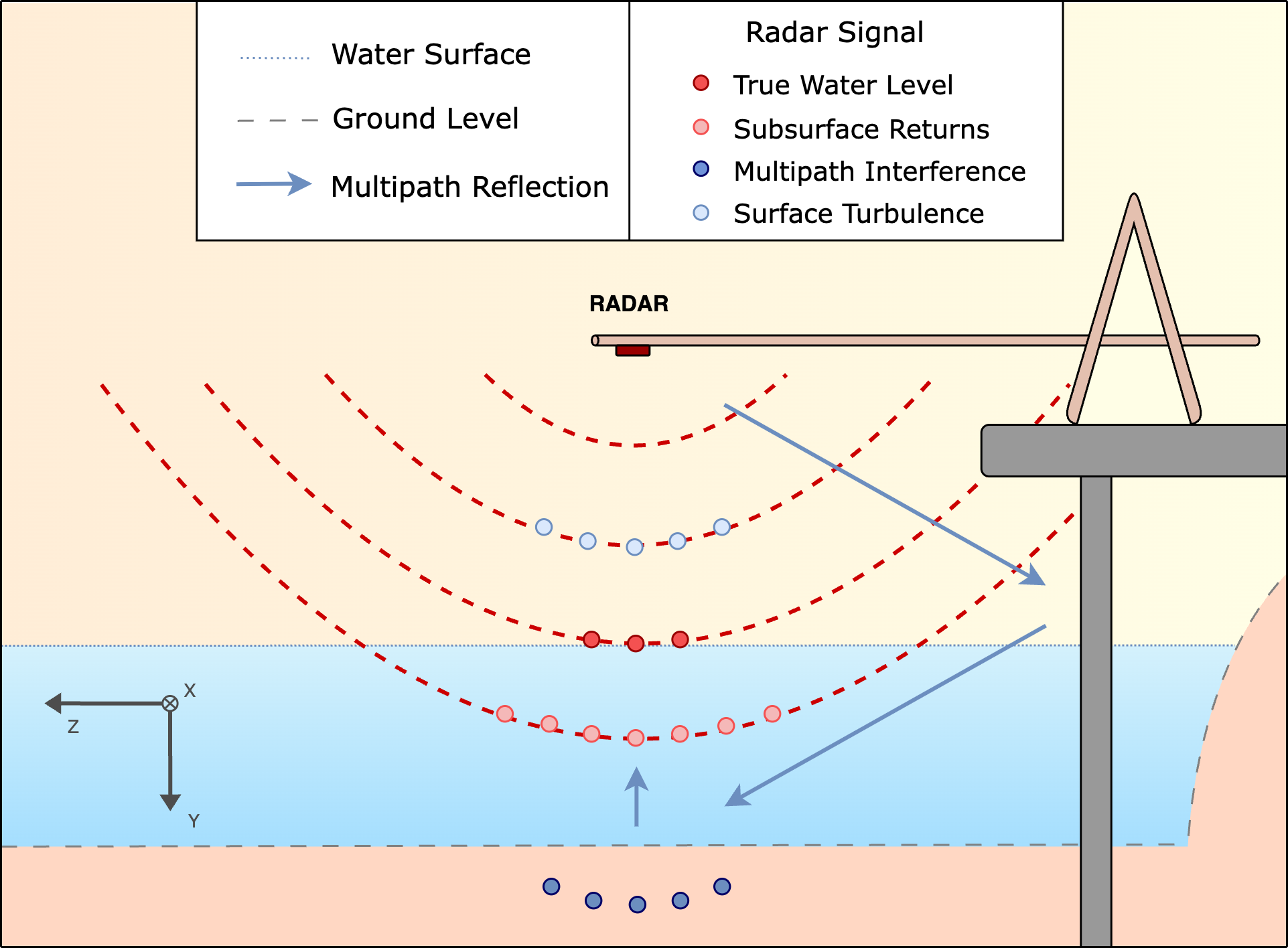}
    \caption{Schematic of the experimental setup. The radar sensor (dark red) is mounted at the end of a fixed horizontal pole and emits radio signals (dashed arcs). The signal can penetrate surfaces such as water (dotted blue) and ground (dashed grey). Reflections from metallic structures may follow indirect paths (blue arrows), producing false returns with large apparent $y$ coordinates (dark blue circles). Additional false returns may arise from sensor noise near the sensor (pale blue circles) or from within the water column (pale red circles). Valid surface returns are shown in bright red circles.}
    \label{fig:rig}
    \vspace{-4em}
\end{wrapfigure}

Figure~\ref{fig:rig} illustrates the sensing rig in a 2D plane and highlights different types of point returns expected during operation. Reflections from the water surface (bright red) may appear alongside returns from the subsurface water mass (pale red) and sensor noise at shorter distances (pale blue). Additionally, radar signals may reflect off surrounding surfaces, creating multipath effects. As shown, waves can reflect from the metallic pier beneath the rig and subsequently from the bottom of the water body, producing spurious point returns at large apparent distances (dark blue). These artifacts arise from signals traversing longer, non-linear propagation paths. Consequently, it is necessary to distinguish true water surface returns from noisy or multipath reflections in order to derive an accurate estimate of the water level.

\subsection{Water Level Estimation}

This work uses straightforward filtering and aggregation techniques to establish the true distance value $d$. To smooth the signal, consecutive measurements are grouped into overlapping windows of size $w$. A point cloud $P_i$ is formed by accumulating $w$ consecutive radar measurements starting from the $i$-th one as:
\begin{equation}
P_i = \left\{ (x, y, p) \in \mathbb{R}^3 \;\middle|\quad
x_{\min} \leq |x| \leq x_{\max},\quad
y_{\min} \leq y \leq y_{\max},\quad
y \geq 0,\quad
p \geq p_{min},\quad
p_{min} \geq 0
\right\},
\end{equation}
\noindent where 
\begin{itemize}
    \item $p$ is the signal intensity at point $(x, y)$,
    \item $p_{min}$ is a filtering parameter corresponding to the minimum signal intensity, and
    \item $x_{\min}$, $x_{\max}$, $y_{\min}$, $y_{\max}$ are filtering parameters that define the region of interest in front of the radar sensor. 
\end{itemize}

Points may be filtered out if they are too close, outside the radar’s center plane, or of low intensity. Then, the distance from the sensor to the water surface is estimated as follows:
\begin{equation}
d_{\text{run}} = \operatorname{mode}\left(F(P_1), F(P_2), \dots, F(P_{N-w+1})\right),
\end{equation}
where:
\begin{itemize}
    \item \( N \) is the total number of measurements in the run,
    \item \( w \) is the aggregation window size,
    \item \( F \) is a function that maps each point cloud to a single representative \( y \)-value (e.g., minimum \( y \), mean \( y \), median \( y \), or the \( y \)-value of the point with maximum intensity \( p \)),
    \item \( \operatorname{mode} \) selects the most frequently occurring value among the $N-w+1$ estimates.
\end{itemize}

\section{Experimental Results}
\subsection{Manual Deployments}

\begin{table}
\centering
  \caption{Sample Manual Deployment Results. $w$ denotes the measurement accumulation window size, with 0 being no sliding window applied. $F$ represents one of the aggregation functions that showed the least error in the run. Two runs with the true water level of 1.13~m and 0.905~m respectively are presented in this table.}
  \label{tab:manual}
  \begin{tabular}[t]{@{}llllll@{}}
    \hline
    \textbf{Sensor} &  $w$ & $F$ & \textbf{True Water Level, m} & \textbf{Measured Water Level, m} & \textbf{Error, m}\\
    \hline
AWR1843 & 2   & max $p$  &  1.13 & 1.12  & 0.01 \\
        & 2   & min $y$  & 0.905 & 0.86  & 0.045 \\

IWR1443 & 0   & min $y$  & 1.13  & 1.14  & 0.01 \\
        &  0  & min $y$  & 0.905 & 0.938 & 0.033 \\

Cascade  & 0 & max $p$ & 1.13  & 1.13 & 0.0 \\
         & 0 & max $p$ & 0.905 & 1.01 & 0.105 \\
    \hline
  \end{tabular}
\end{table}

Two short-term manual deployments are presented to evaluate basic sensor performance under field conditions. Each deployment consisted of a single run lasting several minutes, during which measurements from all three radar sensors were recorded. Although numerous manual deployments were conducted to explore hardware configurations, mounting conditions, and filtering parameters, the two deployments reported here were selected to illustrate the typical performance trends observed across these experiments. This subsection therefore provides representative examples rather than an exhaustive statistical analysis.

In both deployments, the groundtruth water level was measured manually using a marked pole placed vertically beneath the sensor. The true distance from the sensor to the water surface was then computed based on the known mounting height of the sensing rig. The groundtruth water levels of $1.13$~m and $0.905$~m correspond to the first and second deployments, respectively.

\textbf{Table~\ref{tab:manual}} summarizes the results for all three sensors in each deployment. Each row corresponds to a single run from one sensor, for which the estimated water level is compared directly against the known groundtruth. The parameters $w$ and $F$ denote the measurement aggregation window size and the aggregation function, respectively, as defined in Section~III. These values are reported to indicate which filtering configuration yielded the lowest absolute error for each sensor within the run, rather than implying that the parameters were fixed or tuned specifically for these deployments.

Across both deployments, all three sensors were able to detect the water surface within a few centimeters of the groundtruth. However, the Cascade sensor required empirical narrowing of the sensing region to achieve acceptable accuracy, whereas the AWR1843 and IWR1443 achieved comparable performance with less restrictive filtering. These observations suggest that the increased hardware complexity of the Cascade radar does not provide a clear advantage for static water level sensing in this setting, making it unnecessary for low-cost environmental monitoring applications.

The first deployment was additionally used to evaluate tolerance to sensor tilt. The sensor was mounted at inclination angles of 15° and 28°, and the tilt was compensated using IMU-based coordinate transformations. The estimation method maintained accuracy under these moderate angular deviations, indicating robustness to non-ideal mounting conditions.

Based on these results, the AWR1843 and IWR1443 sensors were selected for the subsequent long-term automated deployment.

\subsection{Automated Deployment}
To evaluate long-term sensing performance under more realistic conditions, the AWR1843 and IWR1443 sensors were deployed for a 3-day automated experiment. The rig was programmed to record one run every 30 minutes, producing a total of 114 runs. Each run lasted approximately 3 minutes, with the AWR1843 producing 10 radar measurements per second and the IWR1443 producing 1.6. Unlike in the manual deployments, the groundtruth was not measured manually but instead obtained from a pre-installed pressure-based depth sensor that made a measurement every 15 minutes. For each radar run, the closest groundtruth depth measurement was selected by timestamp alignment. Since the pressure sensor reports depth and the radar measures distance to the surface, the two quantities are inversely related and reported as deltas for consistent comparison.

\begin{table}
\centering
  \caption{Best filtering parameters selected for each sensor based on minimum MSE across all runs. $F$ is the aggregation function, $w$ is the window size, and $\Delta$MSE is the mean squared error between estimated and true water level deltas.}
  \label{tab:final_params}
  \begin{tabular}[hbtp]{@{}llllllll@{}}
    \hline
    \textbf{Sensor} & $F$ & $w$ & $y_{min}, m$ & $x_{max}, m$ & $p_{max}$, \% & $i_{max}$ & $\Delta$MSE, m$^2$ \\
    \hline
    AWR1843  & min $y$   & 5  & 0.1   & –   & 75   & 5 & $1.055 \times 10^{-4}$ \\
    IWR1443  & mean $y$  & 5  & –     & 1   & –    & 25 & $7.8 \times 10^{-6}$ \\
    \hline
  \end{tabular}
\end{table}

\begin{wrapfigure}{r}{0.5\textwidth}
\centering
\vspace{-1em}
\includegraphics[width=\linewidth]{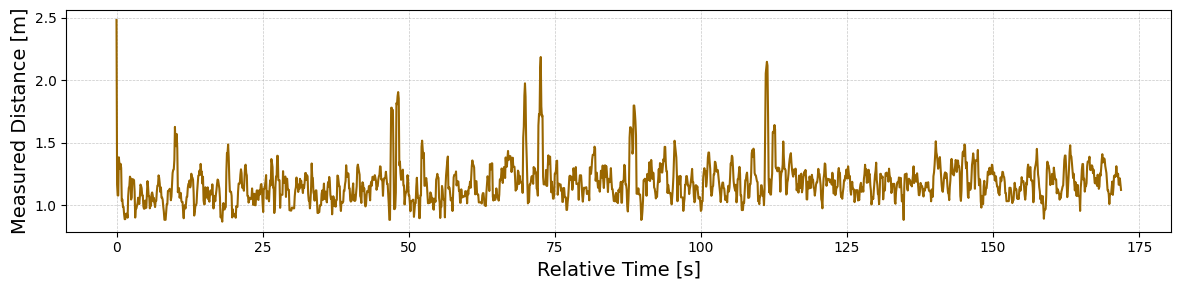}
\vspace{-1.5em}
\caption*{(a) AWR1843}
\vspace{0.5em}
\includegraphics[width=\linewidth]{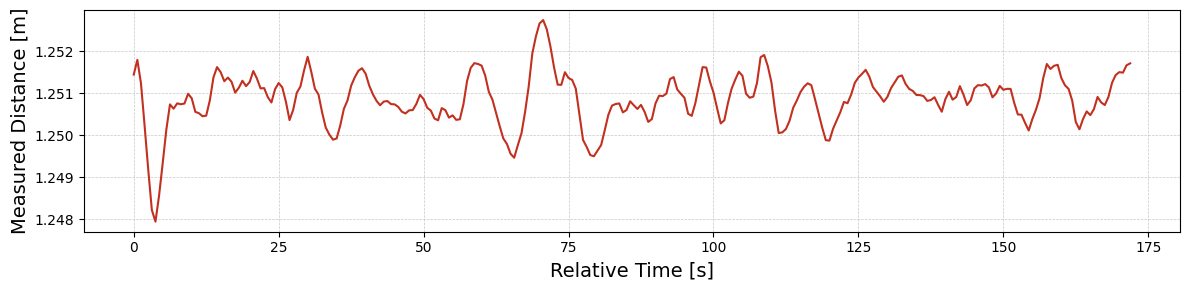}
\vspace{-1.5em}
\caption*{(b) IWR1443}
\caption{Absolute distance-to-surface measurements from a single automated run. The horizontal axis denotes time relative to the start of the run.}
  \label{fig:auto_sample}
\end{wrapfigure}

The difference between the two sensors is illustrated in \textbf{Figure~\ref{fig:auto_sample}} that shows the estimated water levels over time using the first run in the series of 114 as an example. Both sensors were evaluated using their final filtering parameters (\textbf{Table~\ref{tab:final_params}}), except for the aggregation function of the AWR1843, which was changed from minimum $y$ to mean $y$ for visualization purposes. The IWR1443 produces stable estimates across the measurements, while the AWR1843 results appear significantly more noisy.

The automated deployment was evaluated by comparing the estimated water level deltas from each run to the corresponding groundtruth deltas. A grid search over the parameters listed in the Methodology section identified the most effective filtering configuration for each sensor (Table~\ref{tab:final_params}) using the mean squared error (MSE), computed as:
\begin{equation}
\text{MSE} = \frac{1}{N} \sum_{i=1}^{N} \left( \hat{d}_i - d_i \right)^2,
\end{equation}
where \( \hat{d}_i \) is the estimated distance to the water surface for the \(i\)-th run, \( d_i \) is the corresponding groundtruth distance, and \(N\) is the number of runs. The combination that minimized MSE was selected as the final configuration for each sensor. 

\begin{wrapfigure}{r}{0.5\textwidth}
\vspace{-1em}
    \centering
    \includegraphics[width=\linewidth]{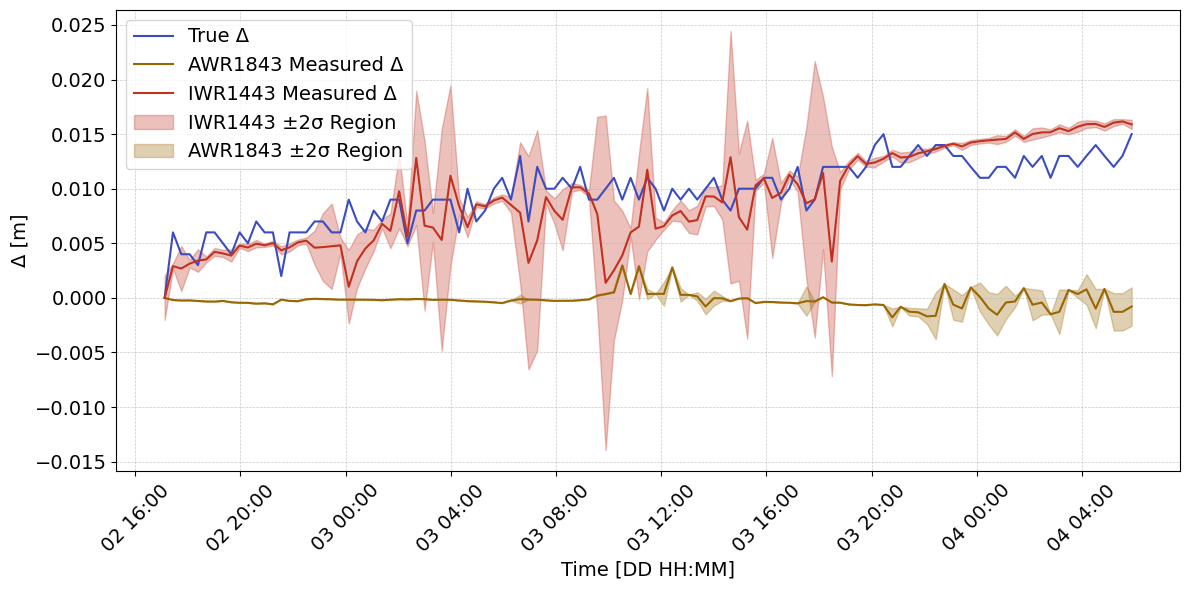}
    \caption{Water level deltas across the entire duration of the automated deployment, computed relative to the initial measurement of each sensor. The distance deltas of the AWR1843 (yellow) and IWR1443 (red) are compared to inverse groundtruth depth deltas (blue).}
    \label{fig:deltas}
    \vspace{-3em}
\end{wrapfigure}

The final sensor configurations are compared in \textbf{Figure~\ref{fig:deltas}}. As the water level gradually dropped during the 3-day deployment, the IWR1443 measurements (red) show a corresponding increase in estimated distance, closely matching the inverse groundtruth depth (blue). The AWR1843 estimates (yellow) remain near zero, as the aggregation function that minimized the overall MSE favored a near-constant estimate under the slowly varying water level conditions of the deployment. Overall, the IWR1443 radar sensor demonstrates the lowest MSE of $7.8 \times 10^{-6}$ m$^2$, corresponding to a root-mean-square error of approximately 2.8~mm.

\section{Discussion}

This work demonstrates that commercially available mmWave radar sensors can achieve sub-centimeter accuracy in water level monitoring under real-world field conditions using lightweight filtering and aggregation. Across both short-term manual experiments and a continuous 3-day automated deployment, accurate water level estimates were obtained without requiring sensor-specific tuning or calibration beyond the use of predefined configuration files.

Among the evaluated sensors, the IWR1443 produced the most accurate and stable measurements in the automated deployment, achieving the lowest mean squared error over the full duration of the experiment. In this deployment, approximately 20 seconds of static observation were sufficient to obtain consistent water level estimates, allowing the sensing process to be performed periodically rather than continuously. This property is beneficial for long-term monitoring scenarios where power consumption or duty cycle may be constrained. It is important to note that the IWR1443 was operated using a waveform configuration provided by the sensor manufacturer and designed specifically for fluid sensing, while the AWR1843 and Cascade sensors were evaluated using their respective default configurations. No parameters within these configurations were modified, and all configuration files were used as provided by the manufacturer. Within this context, the Cascade radar produced competitive results in short-term manual deployments; however, its increased hardware complexity did not yield a clear performance advantage for static water level estimation in this study.

Overall, the results suggest that accurate radar-based water level monitoring can be achieved using compact, low-cost sensors operating with standard configurations and minimal post-processing. The experiments presented here focus on static sensing scenarios, where the sensor remains stationary and oriented toward the water surface. Extending this approach to more complex or non-static settings may require additional mechanisms to distinguish water surfaces from surrounding terrain or other reflective structures. Incorporating a simple water/non-water detection step or surface classification method could improve robustness for robotic platforms operating in unstructured environments.

\section*{Data Availability Statement}
All data, sensor configurations, and code used in this work are openly available at~\cite{zenodo}. No proprietary materials were used.

\medskip

\bibliographystyle{unsrt}
\bibliography{references}{}

@book{intro1book,
  author    = {National Research Council},
  title     = {Hydrologic Hazards Science at the U.S. Geological Survey},
  publisher = {National Academies Press},
  year      = {1999},
  doi       = {10.17226/6385}
}

@article{intro2,
  author  = {Kogan, F. N.},
  title   = {Global drought and flood-watch from NOAA polar-orbiting satellites},
  journal = {Adv. Space Res.},
  volume  = {21},
  number  = {3},
  pages   = {477--480},
  year    = {1998},
  doi     = {10.1016/S0273-1177(97)00883-1}
}

@inproceedings{intro3,
  author    = {Nyborg, L. and Sandholt, I.},
  title     = {NOAA-AVHRR Based Flood Monitoring},
  booktitle = {Proc. IEEE IGARSS},
  year      = {2001},
  pages     = {1696--1698},
  doi       = {10.1109/IGARSS.2001.977041}
}

@article{intro4,
  author  = {Domenikiotis, C. and Loukas, A. and Dalezios, N. R.},
  title   = {The use of NOAA/AVHRR satellite data for monitoring and assessment of forest fires and floods},
  journal = {Nat. Hazards Earth Syst. Sci.},
  volume  = {3},
  number  = {1/2},
  pages   = {115--128},
  year    = {2003},
  doi     = {10.5194/nhess-3-115-2003}
}

@article{harlow2024new,
  author  = {Harlow, K. and Jang, H. and Barfoot, T. D. and Kim, A. and Heckman, C.},
  title   = {A New Wave in Robotics: Survey on Recent MmWave Radar Applications in Robotics},
  journal = {IEEE Trans. Robot.},
  volume  = {40},
  pages   = {4544--4560},
  year    = {2024},
  doi     = {10.1109/tro.2024.3463504}
}

@misc{ajay2023rmap,
  author       = {Mopidevi, A. N. and Harlow, K. and Heckman, C.},
  title        = {RMap: Millimeter-Wave Radar Mapping Through Volumetric Upsampling},
  year         = {2023},
  eprint       = {2310.13188},
  archivePrefix= {arXiv},
  primaryClass = {cs.RO}
}

@misc{kramer2021coloradar,
  author       = {Kramer, A. and Harlow, K. and Williams, C. and Heckman, C.},
  title        = {ColoRadar: The Direct 3D Millimeter Wave Radar Dataset},
  year         = {2021},
  eprint       = {2103.04510},
  archivePrefix= {arXiv},
  primaryClass = {cs.RO}
}

@misc{zenodo,
  author       = {Zavei-Boroda, Anna},
  title        = {Radar Water Measurements 2023},
  month        = jul,
  year         = 2025,
  publisher    = {Zenodo},
  doi          = {10.5281/zenodo.15941467},
  url          = {https://doi.org/10.5281/zenodo.15941467},
}

@article{rw1,
  author  = {Mutschler, Marc A. and Scharf, Philipp A. and Rippl, Patrick, et al.},
  title   = {River Surface Analysis and Characterization Using FMCW Radar},
  journal = {IEEE J. Sel. Top. Appl. Earth Obs. Remote Sens.},
  volume  = {15},
  pages   = {2493--2502},
  year    = {2022},
  doi     = {10.1109/JSTARS.2022.3157469}
}

@article{rw2,
  author  = {Liu, Y. and Wang, H. and Lei, X. and Wang, H.},
  title   = {Real-time forecasting of river water level in urban based on radar rainfall: A case study in Fuzhou City},
  journal = {J. Hydrol.},
  volume  = {603},
  pages   = {126820},
  year    = {2021},
  doi     = {10.1016/j.jhydrol.2021.126820}
}

@article{rw3,
  author  = {Alimenti, F. and Bonafoni, S. and Gallo, E. and Palazzi, V. and Vincenti Gatti, R. and Mezzanotte, P. and Roselli, L. and Zito, D. and Barbetta, S. and Corradini, C. and Termini, D. and Moramarco, T.},
  title   = {Noncontact Measurement of River Surface Velocity and Discharge Estimation With a Low-Cost Doppler Radar Sensor},
  journal = {IEEE Trans. Geosci. Remote Sens.},
  volume  = {58},
  number  = {7},
  pages   = {5195--5207},
  year    = {2020},
  doi     = {10.1109/TGRS.2020.2974185}
}

@article{rw4,
  author  = {Liao, T.-H. and Simard, M. and Denbina, M. and Lamb, M. P.},
  title   = {Monitoring Water Level Change and Seasonal Vegetation Change in the Coastal Wetlands of Louisiana Using L-Band Time-Series},
  journal = {Remote Sens.},
  volume  = {12},
  number  = {15},
  pages   = {2351},
  year    = {2020},
  doi     = {10.3390/rs12152351}
}

@inproceedings{rl1,
author = {Chen, Wenyao and Feng, Yimeng and Cardamis, Mark and Jiang, Cheng and Song, Wei and Ghannoum, Oula and Hu, Wen},
title = {Soil moisture sensing with mmWave radar},
year = {2022},
isbn = {9781450395090},
publisher = {Association for Computing Machinery},
address = {New York, NY, USA},
url = {https://doi.org/10.1145/3555077.3556472},
doi = {10.1145/3555077.3556472},
booktitle = {Proceedings of the 6th ACM Workshop on Millimeter-Wave and Terahertz Networks and Sensing Systems},
pages = {19–24},
numpages = {6},
location = {Sydney, NSW, Australia},
series = {mmNets '22}
}

@article{rl2,
  author       = {Wang, Jianfeng and Chen, Ken and Li, Tianbin and Mou, Li and Jiang, Runyu and Zhang, Hui and Song, Tao},
  title        = {A new method of monitoring slope displacement using millimeter wave radar},
  journal      = {Landslides},
  volume       = {22},
  number       = {5},
  pages        = {1693--1706},
  year         = {2025},
  doi          = {10.1007/s10346-024-02441-3},
  url          = {https://doi.org/10.1007/s10346-024-02441-3},
  issn         = {1612-5118}
}

@article{ti2017radar,
  title={The fundamentals of millimeter wave sensors},
  author={Iovescu, Cesar and Rao, Sandeep},
  journal={Texas Instruments},
  pages={1--8},
  year={2017}
}

@article{ti2020chirp,
  author={Dham, Vivek},
  title={Programming Chirp Parameters in TI Radar Devices (Rev. A)},
  journal={Texas Instruments},
  type = {Application Report},
  number = {SWRA553A},
  year = {2020}
}

@misc{ti2025rwlm,
  author       = {{Texas Instruments}},
  title        = {High Accuracy Level Sensing User Guide},
  howpublished = {\url{https://www.ti.com/tool/RADAR-TOOLBOX}},
  note         = {Accessed: 2025-03-31}
}

@article{cfar,
  author = {Rohling, Hermann},
  title = {Radar CFAR Thresholding in Clutter and Multiple Target Situations},
  journal = {IEEE Transactions on Aerospace and Electronic Systems},
  volume = {AES-19},
  number = {4},
  pages = {608--621},
  year = {1983},
  doi = {10.1109/TAES.1983.309334}
}

\end{document}